 \documentclass[pmlr,twocolumn]{jmlr} 

\usepackage{booktabs}
\usepackage[utf8]{inputenc} 
\usepackage[T1]{fontenc}    
\usepackage{hyperref}       
\usepackage{url}            
\usepackage{nicefrac}       
\usepackage{microtype}      
\usepackage{stmaryrd}
\usepackage{graphicx}
\usepackage{booktabs}       
\usepackage{amsfonts,amsmath,amssymb}       
\usepackage{nicefrac}       
\usepackage{microtype}      
\usepackage[parfill]{parskip}
\usepackage{mathtools}
\usepackage{cases}
\usepackage{comment}
\usepackage{paralist}
\usepackage{algorithmic}
\usepackage{breqn}
\usepackage{algorithm}
\usepackage{color}
\usepackage{appendix}
\usepackage{xspace}
\usepackage{enumitem}
\usepackage{gensymb}
\usepackage[english]{babel}
\usepackage{xcolor}
\usepackage{todonotes}
\usepackage{wrapfig}
\usepackage{bbm}
\usepackage[load-configurations=version-1]{siunitx} 

\theorembodyfont{\upshape}
\theoremheaderfont{\scshape}
\theorempostheader{:}
\theoremsep{\newline}

\jmlrvolume{LEAVE UNSET}
\jmlryear{2020}
\jmlrsubmitted{LEAVE UNSET}
\jmlrpublished{LEAVE UNSET}
\jmlrworkshop{Machine Learning for Health (ML4H) 2020} 
\DeclareMathOperator*{\argmin}{arg\,min}
\DeclareMathOperator*{\argmax}{arg\,max}
\hypersetup{
colorlinks,
linkcolor={blue},
citecolor={blue},
urlcolor={blue}
}

\mathchardef\mhyphen="2D

\newcommand{\vertiii}[1]{{\left\vert\kern-0.25ex\left\vert\kern-0.25ex\left\vert #1
    \right\vert\kern-0.25ex\right\vert\kern-0.25ex\right\vert}}

\newcommand{\train}{\text{train}}

\newcommand{\pool}{\text{pool}}

\usepackage{array}
\newcommand{\PreserveBackslash}[1]{\let\temp=\\#1\let\\=\temp}
\newcolumntype{C}[1]{>{\PreserveBackslash\centering}p{#1}}
\newcolumntype{R}[1]{>{\PreserveBackslash\raggedleft}p{#1}}
\newcolumntype{L}[1]{>{\PreserveBackslash\raggedright}p{#1}}

\title[Confounding Feature Acquisition for Causal Effect Estimation]{Confounding Feature Acquisition for Causal Effect Estimation}

\newcommand*\samethanks[1][\value{footnote}]{\footnotemark[#1]}

\author{
\Name[1,2]{Shirly Wang\thanks{Equal contribution}} \Email{shirly@layer6.ai}\\
\Name[1,2]{Seung Eun Yi\samethanks} \Email{seungeun@layer6.ai}\\
\addr Layer 6 AI, TD Bank Group\\
\Name[3]{Shalmali Joshi}\thanks{Work done while at the Vector Institute} \Email{shalmali@seas.harvard.edu}\\
\addr Harvard University (SEAS) \\
\Name[1,3]{Marzyeh Ghassemi} \Email{marzyeh@cs.toronto.edu}\\
\addr University of Toronto, Vector Institute
}

\begin{document}

\maketitle

\begin{abstract}
Reliable treatment effect estimation from observational data depends on the availability of all confounding information. While much work has targeted treatment effect estimation from observational data, there is relatively little work in the setting of confounding variable missingness, where collecting more information on confounders is often costly or time-consuming.
In this work, we frame this challenge as 
 a problem of feature acquisition of confounding features for causal inference. 
Our goal is to prioritize acquiring values for a fixed and known subset of missing confounders in samples that lead to efficient average treatment effect estimation. 
We propose two acquisition strategies based on i) covariate balancing (\texttt{CB}), and ii) reducing statistical estimation error on observed factual outcome error (\texttt{OE}). 
We compare \texttt{CB} and \texttt{OE} on five common causal effect estimation methods, and demonstrate improved sample efficiency of \texttt{OE} over baseline methods under various settings. We also provide visualizations for further analysis on the difference between our proposed methods.

\end{abstract}

\section{Introduction}
\label{sec:intro}
Reliable causal effect estimation (CEE) from observational data is an important step toward advancing healthcare and science, and much recent work has targeted this problem~\citep{alaa2017bayesian,johansson2016learning,shalit2017estimating}.
However, there are several practical challenges in observational health settings. First, reliable CEE hinges on observing all confounding attributes, e.g., attributes like race that may affect both treatment and outcome.~\citep{pearl2000causality, madras2019fairness, zhang2018fairness} Nevertheless, in many cases not all confounders may be known~\citep{miao2018identifying,pearl2000causality,rubin1974estimating}, making the effect estimate unidentifiable without additional information and/or constraints~\citep{pearl2000causality}. Further, even if all confounders are known, it may be difficult to collect their values due to high costs to the institution, and/or potential loss of confidentiality~\citep{coston2019fair}. For example, while collecting everyday life variables like diet, and physical exercise is costly, they are likely to affect patients with Alzheimer's disease and can influence the evaluation of drug-disease interactions among these patients~\citep{liyanage2018hidden}. When such variables cannot be collected, a proxy model can be built to impute missing values, albeit at the expense of statistical biases and instabilites.~\citep{chen2019fairness}. 

Our work focuses on accurately estimating treatment effect through feature acquisition in the presence of missing confounding attributes. In our case, values for a known and fixed subset of confounding variables are missing in most samples, but a costly mechanism can be deployed to acquire them. This is a relevant setting in healthcare as some observational clinical datasets used for causal studies recruit a cohort, and then acquire additional data for samples in this cohort. For example, the UK Biobank \citep{biobank2014uk} recently collected and released COVID-related data from a subset of their established cohort of participants. Our work is useful for prioritizing the acquisition of data values (from fixed cohorts) that are most beneficial for CEE. Note that our formulation is distinct from latent confounding~\citep{pearl2000causality}, where confounding variables are never observed. It also differs from conventional active learning or feature acquisition~\citep{settles2009active}, where data is sampled for supervised learning.
While past work has tried to address data acquisition for CEE, their focus is to acquire \emph{counterfactual} outcomes for an observed sample (from an expert) under a treatment different from what exists in the observational data (without missingness)~\citep{sundin2019active}. 
In contrast, our work focuses on acquisition of missing values in known confounders as opposed to counterfactual labels to obtain reliable estimates of average treatment effects (ATE)\footnote{See Definition~\ref{def:ATE}.}, when values of some confounding variables is available for only a small subset of samples.

We propose two acquisition strategies based on i) covariate balancing (\texttt{CB}) between treated and control groups, and ii) statistical estimation error to obtain missing values of known confounders for efficient CEE (\texttt{OE}). Using a semi-synthetic dataset, we compare the effectiveness of these strategies in reducing the need for labelling confounding values to estimate accurate ATE. Our experiments involve multiple commonly used CEE models, and multiple simulation settings where information from the missing confounder can be partially recovered through other covariates. 
We analyze the explore-exploit trade-off of both methods, visualizing samples acquired using PCA, and find that \texttt{OE} have more exploratory bias than \texttt{CB}.
We further quantify treated vs control sample preference, finding that \texttt{OE} prefers to acquire samples to minimize errors for the control group as they have a more complex response function, while \texttt{CB} maintains a more balanced feature acquisition ratio over groups.

Our contributions are summarized as follows\footnote{Code at \url{https://github.com/MLforHealth/confounder-acquisition}}:
\begin{itemize}[leftmargin=*]
    \item We propose two feature acquisition strategies to acquire missing values of known confounders for efficient CEE. Both strategies are compatible with common CEE models.
    \item We demonstrate that when the missing feature is independent of other covariates, \texttt{OE} achieves the best sample efficiency when used with most CEE models, and \texttt{CB} offers some initial benefits in reducing ATE errors. The benefits of \texttt{OE} persist even when the missing feature becomes more correlated with other covariates.
    \item We provide insights into the impact of different strategies on samples acquired and show that focusing on statistical estimation error prioritizes early exploration during acquisition and adapts to different complexities of the outcome generating function for the treated and control group.
\end{itemize}

\section{Related Work}
Reliable CEE from observational data is a classical statistical challenge addressed under two main frameworks of potential outcomes and causal graphical models~\citep{pearl2000causality,rubin1974estimating}. Recent works in modern machine learning focus on CEE using parametric assumptions or balancing approaches to improve effect estimates from observational data \citep{alaa2017bayesian,johansson2016learning,shalit2017estimating}. A majority of such methods are developed assuming no hidden confounding or relaxing them under certain conditions~\citep{louizos2017causal,wang2019blessings,miao2018identifying}. 
However, very few methods have considered the problem of missingness in known confounding variables from the perspective of data acquisition or active learning.

Data acquisition in machine learning is designed for supervised learning to obtain target labels, with common strategies including query by committee~\citep{freund1997selective}, uncertainty sampling~\citep{lewis1994sequential}, and information-based loss functions~\citep{settles2009active}. Data acquisition of features has also been explored for supervised learning~\citep{melville2005expected,saar2009active,shim2018joint,janisch2020classification}. However, applicability of such methods for acquiring pre-treatment confounding variables is not well studied. Particularly, while motivations in supervised learning are simply to fit the associational distribution accurately, CEE is invalid without controlling for all confounding. 

In the context of causality, active learning is primarily used for experiment design for causal discovery and obtaining interventional information \citep{he2008active, yan2019label}. 
In our work, we assume that the underlying causal dependencies are already known (and correspond to Figure~\ref{fig:graph}) and our goal is to improve statistical estimates of causal effects by strategically acquiring costly but missing confounding features without which estimation is impossible. 

\section{Methodology}
We build on the potential outcomes framework of~\citet{rubin1974estimating} for causal inference from observational data. We formalize our setup in the following.

\subsection{Setup}
\begin{figure}[htbp!]
    \centering
    \includegraphics[width=\linewidth/2]{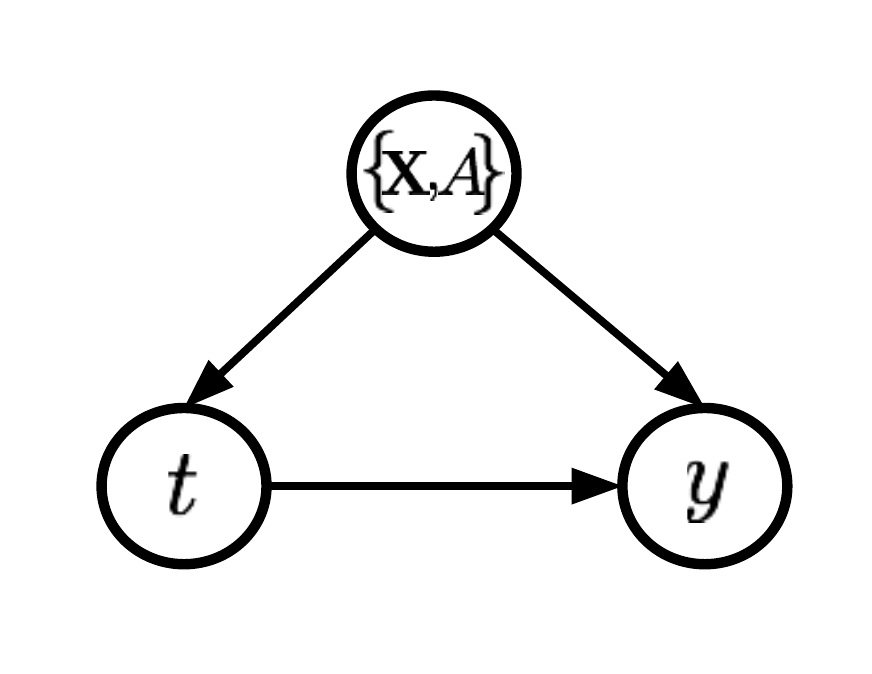}
    \caption{\small{Causal graph of observational data considered in our setting.}}
    \label{fig:graph}
\end{figure}
We consider a setting where we have access to observational data with base covariates $X$, missing attribute $A$, treatment $t$, and outcome $y$ (see Figure~\ref{fig:graph} for the causal diagram assumed throughout this work). 
Both $A$ and $t$ are considered binary variables throughout the paper for convenience. However, the proposed methods also work if the missing attribute is categorical and can be generalized in the presence of multiple missing attributes. $y_0$ is the potential outcome for this sample if $t=0$ and $y_1$ is the outcome corresponding to $t=1$. For each independent and identically distributed sample ($X$, $A$, $t$), we can only observe one potential outcome depending on the treatment assignment: $y = y_0$ if $t=0$, and $y=y_1$ if $t=1$. We assume the following throughout this work:

\begin{assumption}[Ignorability] \label{as:cond_ind}\\
Potential outcomes are independent of the treatment given covariates $\{X, A\}$:
\begin{equation*}
    y_{1}, y_{0} \, \perp \!\!\! \perp  \, t \, | \, X, A
\end{equation*}
\end{assumption}

\begin{assumption}[Common Support]\label{as:comm_supp} \\
$0< p(t=1 | X,A) < 1$ for all values of $X,A$ with $p(X,A) > 0$
\end{assumption}

In addition, $A$ is \textit{missing not at random} (MNAR) (i.e. the missingness of $A$ depends on its actual values, see Appendix \ref{app:sim_algo}). We call $\text{D}_\pool$ the set of samples with missing $A$. This hinders an accurate estimation of the average treatment effect (ATE, Definition~\ref{def:ATE}) $E[y_1 - y_0]$. Our goal is to strategically query samples from $\text{D}_\pool$ to obtain actual values of $A$ so that the treatment effect estimate can be improved. 

\begin{figure}[thbp!]
    \centering
    \includegraphics[width=\linewidth]{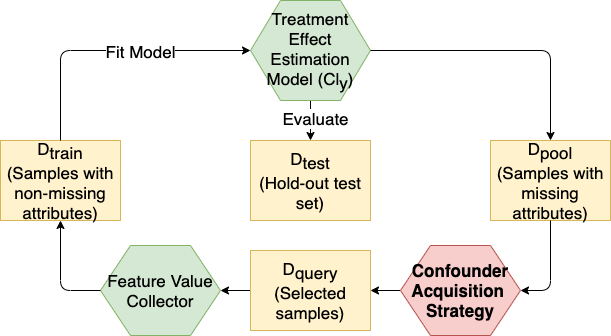}
    \caption{\small{Pipeline for data acquisition for CEE. The focus of our work is on confounding feature acquisition (pink block).}}
    \label{fig:pipeline}
\end{figure}

We use an iterative procedure to evaluate the effectiveness of our proposed strategies. Missing features are acquired in batches and used in subsequent model training to obtain effect sizes. This procedure is summarized in the pipeline Figure \ref{fig:pipeline} and Algorithm \ref{alg:main_procedure}. 

 \begin{algorithm}[htbp!]
 \caption{Feature acquisition for CEE}
  \label{alg:main_procedure}
 \begin{algorithmic}[1]
 \STATE {\textbf{Input:} $\text{D}_{\text{train}}, \text{D}_{\text{test}}, \text{D}_{\text{\pool}}$, Models $\texttt{Cl}_y$, $\texttt{Cl}_{A}$, Acquisition Method (\texttt{CB} or \texttt{OE}), Batch size: $\beta$, ATE variance threshold (hyperparameter): $\sigma_{ate}^2$.}
 \WHILE{$|\text{D}_{\pool}| > 0$ and $Var(\epsilon_{ATE}) > \sigma_{ate}^2$}
     \STATE $\texttt{Cl}_y \leftarrow \texttt{Cl}_y (X_{\train}, A_{\train}, t_{\train})$
     \STATE $\texttt{Cl}_A \leftarrow \texttt{Cl}_A (X_{\train}, t_{\train})$
     \STATE Choose batch of size $\beta$, i.e. $(X^*, t^*, y^*)_{\beta}$  using Eq.~\ref{eq:CB} for \texttt{CB} or Eq.~\ref{eq:OE} for \texttt{OE} \\
     \STATE $A^*_{\beta} \leftarrow \texttt{Oracle}((X^*, t^*, y^*)_{\beta})$  \\
     \STATE $\text{D}_{\text{train}} \leftarrow \text{D}_{\text{train}} \cup (X^*, t^*, y^*, A^*)_{\beta}$ \\
     \STATE $\text{D}_{\text{pool}} \leftarrow \text{D}_{\text{pool}} \setminus (X^*, t^*, y^*)_{\beta}$ \\
\ENDWHILE
 \RETURN{} $\epsilon_{PEHE}, \epsilon_{ATE}$
\end{algorithmic}
 \end{algorithm}

\subsection{Feature Acquisition Strategies}
We propose two feature acquisition strategies. The first consists of explicitly characterizing and fixing covariate imbalance between treated and control samples. 
Addressing such imbalances is critical for unbiased CEE. The second is motivated by improving estimation error of outcomes for treated and control populations. That is, we estimate the expected utility of decrease in estimation error by acquiring missing confounding value of $A$, and select samples entirely based on statistical errors in potential outcomes estimation. Our methods are described in the following.

\paragraph{Covariate Balancing (CB)}
The presence of confounding creates unwanted dependencies between the treatment assignment and outcome of interest, and is a deterrent to reliable CEE. Adjusting for confounding aims to account for distributional differences between the treatment and control group (i.e., \emph{imbalance}), and is a key element in CEE~\citep{rubin2005causal,shalit2017estimating}.
Our first proposed strategy aims to use feature acquisition to decrease imbalance. We acquire samples from $\text{D}_{\pool}$ that, when added to $\text{D}_{\train}$, reduce the expected Maximum Mean Discrepancy (MMD)\footnote{See Def~\ref{def:MMD}.} between treated and control groups. We denote the population MMD estimated from samples $\texttt{U} \sim P$ and samples $\texttt{V} \sim Q$ by MMD(\texttt{U}, \texttt{V}). 

\begin{small}
\begin{definition}[MMD]\label{def:MMD}
 \begin{equation}\label{eq:mmd}
 \begin{footnotesize}
 \begin{aligned}
     \text{MMD}(P, Q) &= \sup_{\|f\|_{\mathcal{H}} \leq 1 } E_{X \sim P} [f(X)]  
       - E_{X \sim Q} [f(X)] 
    \end{aligned}
    \end{footnotesize}
    \end{equation}
    where $\mathcal{H}$ is the Reproducing Kernel Hilbert Space (RKHS). 
\end{definition}
\end{small}

Since values of $A$ are missing, for each sample in $\text{D}_{\text{pool}}$, we first model $p(A|X, T)$ (using $Cl_A$ learnt with samples in $\text{D}_{\train}$), compute the expected MMD, and choose samples that will minimize the following estimate. Let $\texttt{T}$ be the treated samples in $\text{D}_{\train}$, and $\texttt{C}$ be the control samples in $\text{D}_{\train}$.
\begin{equation}\label{eq:CB}
\begin{footnotesize}
\begin{aligned}
     \underset{(X,t,y) \in \text{D}_{\text{pool}}}{\argmin}
     & E_{A \sim p(A | X, t)}  \big( \llbracket t=1\rrbracket \text{MMD}(\texttt{T} \cup (X, A), \texttt{C}) + \\
     &\, \llbracket t=0 \rrbracket \text{MMD}(\texttt{T}, \texttt{C}  \cup (X, A)) \big)
\end{aligned}
\end{footnotesize}
\end{equation}
where $\llbracket \cdot \rrbracket$ denotes the indicator function. Note that we model $A$ only using other confounding attributes and treatment (not outcome $y$) so that treated and control units are balanced without observing the factual outcome.


\paragraph{Outcome Error (OE)}
Our second strategy is to focus only on overall estimation errors and acquire data that improve estimation directly.  
For each sample in $\text{D}_{\text{pool}}$, we predict the expected outcome $E_{A \sim p(A| X, t)} [\hat{y} | X, A, T]$, where the weights are $p(A|X,T)$. $\texttt{Cl}_{y} \triangleq p(y | X, A, t)$ is the estimator trained on $\text{D}_{\text{train}}$ and $\texttt{Cl}_{A} \triangleq p(A|X,t)$ is trained as in the previous method.
Since the observed factual outcome $y$ is known, the samples that lead to the highest error between the predicted outcome and the observed outcome are picked and the information of $A$ is queried. In other words, we choose samples using:
\begin{equation}\label{eq:OE}
     \underset{(X,t,y) \in \text{D}_{\text{pool}}}{\argmax} |E_{A \sim p(A| X, t)} [\hat{y} | X, A, t] - y|
\end{equation}

While this is closer in spirit to traditional active learning, note that the latter focuses on acquiring the label $Y$ rather than a confounding feature $A$ with the explicit purpose of effect size estimation. Additionally, the source of the statistical errors in outcome estimates comes from confounding variables.

We compare both proposed strategies to two commonly used active learning baseline strategies:
\begin{itemize}[leftmargin=*]
    \item \textbf{Random} This method consists of taking random samples and querying their $A$ values.
    \item \textbf{Uncertainty of $A$ (Uncertainty)} At each iteration, a classifier ($\texttt{Cl}_A$) is trained to predict $A$ from the covariates $X$ and the treatment $t$. Samples from $\text{D}_{\pool}$ with the highest uncertainty on the value of $A$ are selected.
\end{itemize}

\subsection{Models}
We use a random forest classifier \citep{breiman2001random} to train $\texttt{Cl}_A$, used in all strategies to predict $A$ from covariates $X$ and treatment $t$ at each iteration. As for $\texttt{Cl}_y$, we test five different CEE models.  
They range from traditional statistical approaches to more recent machine learning approaches. These models are:
\begin{itemize}[leftmargin=*]
    \item \textbf{Causal Forest (CF)}\citep{wager2018estimation}. CF is an extension of the random forest algorithm adapted to infer causal effects. As treatment effect is directly predicted in the leaf nodes without prediction of factual or counterfactual outcome, \texttt{OE} strategy does not work with CF. 
    \item \textbf{Doubly Robust Estimation (DR)} \citep{bang2005doubly}. DR combines the capacity of linear model and covariate balance from propensity scoring. DR originally combines a logistic regression for propensity scoring and a linear regression for final outcome prediction. We implement a more complex estimator by replacing linear models with two single-layer neural networks. 
    \item \textbf{Multiheaded Multi-Layer Perceptron (MLP\_Multi)}. We fit two Multilayer Perceptron neural networks \citep{sklearn_api}, one for the treated group and the other for the control group to estimate potential outcomes. The final treatment effect estimate is the difference between them.
    \item \textbf{Multiheaded Gaussian Processes (GP\_Multi)}. We fit two Gaussian process regressors with RBF kernels \citep{sklearn_api}, one for the treated group and the other for the control group to estimate potential outcomes. Final treatment effect estimate is taken as the difference between them.
    \item \textbf{Causal Multitask Gaussian Processes (CMGP)}. We augment the CMGP procedure introduced by~\citet{alaa2017bayesian} with our acquisition strategies to estimate treatment effect.
\end{itemize}

\subsection{Data}
We use a semi-synthetic benchmark dataset, Infant Health and Development Program (IHDP) dataset~\citep{hill2007bayesian}, to evaluate the proposed feature acquisition strategies. As in other work, we remove a subset of the population to create a biased dataset \citep{shalit2017estimating,madras2019fairness,johansson2016learning} and use mother's ethnicity $momwhite$ as $A$. 

To ensure that information from $A$ is truly unavailable for most samples, we create independence between other covariates and $A$ by randomly permuting values for $A$. (We later relax this independence assumption in Section \ref{sec:a_exp}).  After normalizing the dataset, we adapt the generation of outcomes using the response surface type B of \citet{hill2007bayesian}. Specifically, we sample the treatments under Bernoulli distributions and potential outcomes under normal distribution. To create a partially observed dataset, we mask 95\% of $A$ values, where the probability of missing depends on unmasked $A$ values. 

Details on the simulation steps are described in Algorithms \ref{alg:algo_a} - \ref{alg:algo_outcomes} in Appendix \ref{app:sim_algo}.

\subsection{Evaluation}
We evaluate the effectiveness of our proposed acquisition strategies in terms of average treatment effect (ATE) and individual treatment effect (ITE) using a hold-out test set. 

For ATE, we measure sample estimate of average absolute error in ATE. For ITE, we measure sample estimates of Precision in Estimation of Heterogeneous Effect (PEHE). 

Let $\hat{y}_{1}(X, A)$ and $\hat{y}_0(X, A)$ be the potential outcomes estimated under any of the aforementioned hypothesis classes. Definition of these metrics are listed below:

\begin{small}
\begin{definition}[Sample Error in ATE]\label{def:ATE}
 \begin{equation}\label{eq:ate}
 \begin{aligned}
 \small
    \epsilon_{ATE} =&  |\frac{1}{n}\sum_{i=1}^n(\hat{y_1}(X_i, A_i)-\hat{y_0}(X_i, A_i)) \\
    -& \frac{1}{n}\sum_{i=1}^n(y_1(X_i, A_i)-y_0(X_i, A_i))|
    \end{aligned}
\end{equation}
\end{definition}

\begin{definition}[Sample PEHE]\label{def:PEHE}
 \begin{equation}\label{eq:pehe}
 \begin{aligned}
     \epsilon_{PEHE} =& \frac{1}{n}\sum_{i=1}^n(\hat{y_1}(X_i, A_i)-\hat{y_0}(X_i, A_i)\\ 
      &\, - (y_1(X_i, A_i)-y_0(X_i, A_i)))^2
    \end{aligned}
    \end{equation}

\end{definition}
\end{small}

Note also that we use the \textit{noiseless} outcome in data generation as $y_1(X_i, A_i)$ and $y_0(X_i, A_i)$. This is consistent to other papers using the IHDP dataset \citep{hill2007bayesian, shalit2017estimating}.

\begin{table*}[h!]
\small
\centering
\setlength{\tabcolsep}{4pt} 
\renewcommand{\arraystretch}{0.5}
\floatconts
    {tab:indep_results}
    {\caption{\small{Comparison of results for different model types and acquisition strategies. 
    Lower is better. $\pm$ shows the width of 95\% confidence interval, running on 500 realizations of simulated IHDP data. Statistically significant better performing strategies for each model are bolded.}}} 
    {\subtable[Optimal performance of each ATE estimation method when all feature values are acquired and all samples are used for training.]{
        \label{tab:opt_performance}
        \begin{tabular}{|p{0.2\linewidth}|%
                        *{5}{R{0.14\linewidth}}|}
        \toprule
                        & \multicolumn{1}{c}{DR} & \multicolumn{1}{c}{CMGP} & \multicolumn{1}{c}{GP\_Multi}  &\multicolumn{1}{c}{MLP\_Multi} & \multicolumn{1}{c|}{CF}  \\
        \midrule
        Optimal $\epsilon_{ATE}$     & 0.65 $\pm$ 0.07   & 0.45 $\pm$ 0.05 & 0.72 $\pm$ 0.07    & 0.81 $\pm$ 0.08    & 0.90 $\pm$ 0.09  \\
        Optimal $\sqrt{\epsilon_{PEHE}}$    & 5.41 $\pm$ 0.42   & 4.44 $\pm$ 0.35 & 7.03 $\pm$ 0.48    & 7.90 $\pm$ 0.58    & 8.28 $\pm$ 0.57        \\
        \bottomrule                                                         
        \end{tabular}}
    \subtable[Average number of samples needed to reach within 1\% of optimal performance.]{
        \label{tab:samples_main}
        \begin{tabular}{|p{0.2\linewidth}|%
                        *{5}{R{0.14\linewidth}}|}
        \toprule
                        & \multicolumn{1}{c}{DR} & \multicolumn{1}{c}{CMGP} & \multicolumn{1}{c}{GP\_Multi}  &\multicolumn{1}{c}{MLP\_Multi} & \multicolumn{1}{c|}{CF}  \\
        \midrule
        \multicolumn{6}{|c|}{Number of Samples Needed to Reach within 1\% of Optimal $\epsilon_{ATE}$}   \\
        \midrule
        Random                  & 168 $\pm$ 13           &165 $\pm$ 14            & 174 $\pm$ 15            & 107 $\pm$ 10      & 136 $\pm$ 13 \\
        Uncertainty             & 160 $\pm$ 13           &154 $\pm$ 14            & 175 $\pm$ 16            & 108 $\pm$ 11      & 129 $\pm$ 13 \\
        CB                      & 162 $\pm$ 13           &137 $\pm$ 13            & 128 $\pm$ 13            & 98 $\pm$ 11       & 108 $\pm$ 12 \\
        OE                      & \textbf{92 $\pm$ 9}    &\textbf{111 $\pm$ 11}   & \textbf{99 $\pm$ 9}     & 117 $\pm$ 10      & \\
        \midrule
        \multicolumn{6}{|c|}{Number of Samples Needed to Reach within 1\% of Optimal $\sqrt{\epsilon_{PEHE}}$}               \\
        \midrule
        Random                  & 451 $\pm$ 12           &487 $\pm$ 9          & 492 $\pm$ 9           & 435 $\pm$ 16       & 425 $\pm$ 13  \\
        Uncertainty             & 422 $\pm$ 12           &478 $\pm$ 10         & 486 $\pm$ 10          & 427 $\pm$ 16       & 427 $\pm$ 13  \\
        CB                      & 413 $\pm$ 12           &472 $\pm$ 10         & 496 $\pm$ 9           & 428 $\pm$ 16       & 434 $\pm$ 13  \\
        OE                      & \textbf{170 $\pm$ 12}  &\textbf{352 $\pm$ 14}& \textbf{398 $\pm$ 9}  & \textbf{343 $\pm$ 18}  & \\           
        \bottomrule                                                         
        \end{tabular}}}
\end{table*}

\section{Results} \label{sec:results}

We measure the effectiveness of different strategies by their empirical sample efficiency in reaching the effect estimate when all samples are acquired or there is no missingness. 
Table \ref{tab:samples_main} summarizes this information in terms of the number of samples needed to reach within 1\% of the optimal performance (presented in Table \ref{tab:opt_performance}). It should also be noted that here we focus on ATE, not ITE, as our acquisition setup is not designed for individual level CEE. We show PEHE for completeness. 

As shown in the table, \texttt{OE} is significantly more effective for DR, CMGP, and GP\_Multi in reaching both optimal error in ATE and optimal PEHE. \texttt{OE} also provides significant benefit in reaching optimal PEHE for MLP\_Multi.  
\begin{figure}[htbp!]
    \centering
    \floatconts{fig:indep_results}
    {\caption{\small{Error in ATE and PEHE. X-axis is the number of points in the queried pool. Each line shows the change in average value of performance metrics in 500 realizations, and the shaded region is the corresponding 95\% confidence interval. Lower is better. Spikes in CMGP results are due to numerical instabilities as a result of implementing CMGP using the GPy package.}}}
    {\includegraphics[width=\linewidth]{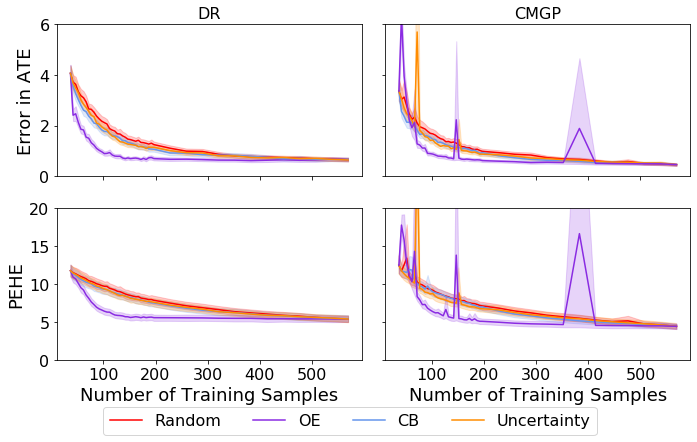}}
\end{figure}

As DR and CMGP are the best models in terms of optimal error in ATE and PEHE, we plot the changes in values of these metrics when running different acquisition strategies with them in Figure \ref{fig:indep_results}. The figures clearly demonstrate the significant benefit from using \texttt{OE}. Additionally, when using DR, \texttt{OE} offers consistent better performance regardless of number of samples queried, as compared to \texttt{CB} and baseline strategies. 
It should be noted that by focusing on a specific source of bias - imbalance - in effect estimates, \texttt{CB} provides some benefits initially for both DR and CMGP but these benefits are generally outweighed by relying on cumulative sources of statistical errors (\texttt{OE}) for feature acquisition.

\section{Analysis}
\subsection{Analysis of Relationship between $A$ and Other Confounders} \label{sec:a_exp}
\begin{table*}[h!]
\setlength{\tabcolsep}{4pt} 
\renewcommand{\arraystretch}{0.5}
\small
\floatconts
 {tab:a_exp_ate}
 {\caption{\small{Sample efficiency to reach optimal errors in ATE while relaxing the independence assumption. 
 Values are the average number of samples needed in training to reach within 1\% of the optimal performance level, with $\pm$ showing the width of 95\% confidence interval, over 100 realizations of the simulated data. Lower is better. Statistically significant better performing strategies for each experiment type are bolded.}}}
 {%
   \subtable[Experiments with noise added to original $A$ values. Columns represent fraction of original $A$ values retained.]{%
    \label{tab:a_exp_original_a_ate}%
    \begin{tabular}{|l|cccccc|}
    \toprule
                         & 0              & 0.2            & 0.4                           & 0.6            & 0.8             & 1              \\
    \midrule
    Random               & 156 $\pm$ 29          & 175 $\pm$ 30         & 161 $\pm$ 27                        & 158 $\pm$ 27         & 160 $\pm$ 29          & 145 $\pm$ 23        \\
    Uncertainty          & 149 $\pm$ 28         & 182 $\pm$ 33         & 128 $\pm$ 22                        & 142 $\pm$ 26        & 123 $\pm$ 23          & 119 $\pm$ 24         \\
    CB                   & 144 $\pm$ 26         & 144 $\pm$ 25        & 149 $\pm$ 24                      & 122 $\pm$ 24         & 125 $\pm$ 25          & 150 $\pm$ 31         \\
    OE                   & \textbf{87 $\pm$ 19} & \textbf{81 $\pm$ 18} & 91 $\pm$ 20                & 84 $\pm$ 18 & 108 $\pm$ 26 & 80 $\pm$ 16 \\
    \bottomrule
    \end{tabular}
   }\\
   \subtable[Experiments with Bivariate gaussian simulation. Columns represent correlation coefficient between simulated $A$ and $birthweight$.]{%
    \label{tab:a_exp_birthweight_a_ate}%
    \begin{tabular}{|l|cccccc|}
    \toprule
                         & 0              & 0.2            & 0.4                           & 0.6            & 0.8             & 1              \\
    \midrule
    Random               & 193 $\pm$ 33         & 171 $\pm$ 28          & 182 $\pm$ 32                        & 159 $\pm$ 30         & 140 $\pm$ 28          & 141 $\pm$ 25         \\
    Uncertainty          & 156 $\pm$ 30         & 143 $\pm$ 27         & 139 $\pm$ 28	                       & 142 $\pm$ 31         & 109 $\pm$ 19          & 134 $\pm$ 24        \\
    CB                   & 134 $\pm$ 23         & 184 $\pm$ 32         & 154 $\pm$ 28                       & 142 $\pm$ 25         & 143 $\pm$ 27          & 142 $\pm$ 28         \\
    OE                   & 97 $\pm$ 23 & \textbf{76 $\pm$ 16} & \textbf{92 $\pm$ 19}                & \textbf{87 $\pm$ 18} & 87 $\pm$ 18 & \textbf{81 $\pm$ 15} \\
    \bottomrule
    \end{tabular}
   }
 }
\end{table*}
In this section, we evaluate whether the benefits of our proposed methods persist even when the missing attribute is correlated with other confounders. DR is used as the CEE model for all experiments in this section as it is the one of the best performing models and requires a significantly shorter run time compared to CMGP. We run 100 realizations of simulated IHDP data, and vary dependency in two ways:

\paragraph{Original $A$ Values}
We use the original values of $momwhite$ in the IHDP dataset. At one extreme, we replace $momwhite$ values in all samples with randomly generated values (i.e. the setting that produces our main results in Section \ref{sec:results}). At the other extreme, we keep the original $momwhite$ values that are correlated with other covariates naturally. In between, we randomly replace $20\%$, $40\%$, $60\%,$ or $80\%$ of original $A$ values with a random value. When $momwhite$ value is generated, it follows a Bernoulli distribution where the positive probability is equal to the sample probability in the original dataset. This evaluates benefits of each strategy with varying levels of noise affecting model fit for $p(A|X,t)$.

\paragraph{Bivariate Gaussian Simulation}
We simulate $momwhite$ such that it has varying degrees of correlation with $birthweight$ denoted by $X_b \subseteq X$ (one of the covariates in IHDP). More specifically, we first generate $A$ such that $[logit(A),X_b]$ is jointly Gaussian, with $logit(A)$ following a standard normal distribution, and $X_b$ following a normal distribution with mean and standard deviation matching those of $birthweight$. Correlation coefficient between $logit(A)$ and $X_b$ varies between $0$ and $1$. $A$ is generated by thresholding $logit(A)$ such that the proportions of $momwhite$ is the same as that in the original dataset. This evaluation assesses whether benefits of using \texttt{CB} and \texttt{OE} persist at different levels of correlation, when we explicitly model the relationship between $A$ and other covariates $X \supseteq X_{b}$. 


Tables \ref{tab:a_exp_ate} summarizes the number of samples needed under each acquisition strategy to reduce ATE to a level that is within 1\% of the optimal performance. As shown in the table, \texttt{OE} consistently outperforms the baseline strategies despite varying level of correlation between $A$ and other covariates. We also observe that as $A$ becomes more correlated with other covariates, it generally requires fewer acquired samples to reach optimal performance. Evaluation of PEHE gives similar patterns and is presented in Appendix \ref{app:a_exp_pehe}.

\begin{figure*}[tb]
\small
\floatconts
  {fig:vis_all}
  {\caption{\small{Comparison of \texttt{CB} and \texttt{OE} strategies using CMGP as the CEE model.}}}
  {
    \subfigure[Visualization of training samples using first two principal component of all covariates (including $A$) as more feature values are acquired. Results shown are obtained on a single realization of simulated IHDP data. Results on other realizations show the same pattern.][c]{\label{fig:vis}%
      \includegraphics[width=0.8\linewidth]{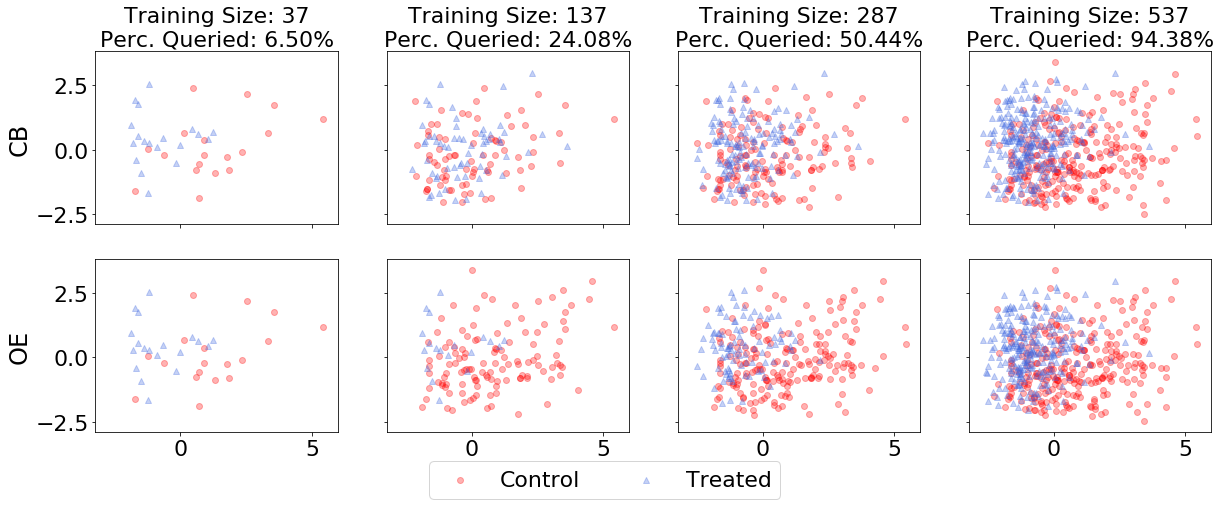}}%
    \\
    \subfigure[Average number of treated and control observations as more feature values are acquired, across 500 realizations of simulated IHDP data.][c]{\label{fig:vis_segment}%
      \includegraphics[width=0.8\linewidth]{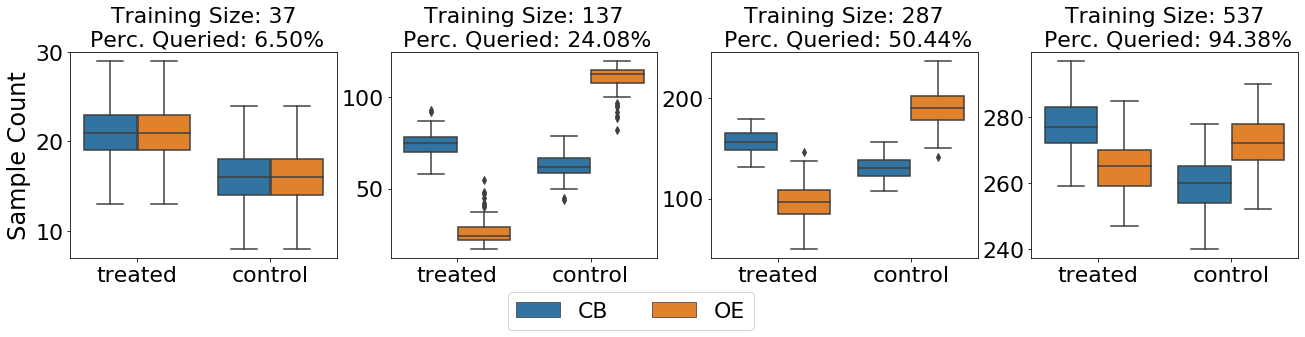}}
  }
\end{figure*}

\subsection{Comparison of Acquisition Strategies}
We analyze the performance difference between \texttt{CB} and \texttt{OE} by visualizing what samples are prioritized by each strategy, as well as comparing some resulting summary statistics of the acquired samples. Results shown in this section use CMGP for $\texttt{Cl}_y$ because of its strong performance and are under the same setting as our main experiment (i.e. $A$ is independent of other covariates). We have two findings:


\paragraph{Explore-Exploit Tradeoff} One challenge in a traditional active learning problem is how explore-exploit trade-off is handled~\citep{bondu2010exploration}. To understand how our two strategies handle this issue, we assess the imbalance at every stage of our procedure by visualizing the first two principal components of all pre-treatment confounders $(X, A)$. Figure \ref{fig:vis} shows a 2D presentation of samples in the initial labeled set, as well as when samples are acquired (out of a total of $569$). While \texttt{CB} first acquires points similar to the already queried point, \texttt{OE} explores more and acquires points from all possible regions in the input space. This is especially true for samples in the control group, where the representation is less clustered. When most samples are acquired, however, the input space is well explored and represented in the training samples by both strategies. Our results in Section \ref{sec:results} suggest that with this particular setting, the benefit of early exploration in \texttt{OE} outweighs the benefit of early exploitation in \texttt{CB}.

\paragraph{Treatment vs Control Samples} We next look at how \texttt{CB} and \texttt{OE} differ in terms of number of treated and control samples acquired. As shown in Figure \ref{fig:vis_segment}, \texttt{OE} strategy queries more points in the control group in the early stage. This is because the expected outcome for the control group is an exponential function of the covariates and has a larger range as compared to the mean outcome for the treated group. \texttt{OE} strategy learns that it is more difficult to predict the a control outcome and therefore queries more control samples. On the other hand, as \texttt{CB} strategy focuses on reducing imbalance, it maintains a more balanced treatment ratio in the acquired samples. \texttt{OE} strategy has a performance advantage in this setting as the outcome generating processes have different complexities between treatment and control group. This advantage of \texttt{OE} empirically outweighs imbalance issues and is an interesting statistical trade-off that should be explored for feature acquisition strategies intended for CEE. 


\section{Conclusion}

Confounding feature acquisition strategies for CEE and down-stream decision making need special attention, particularly in healthcare, where sensitive and/or costly but critical pre-treatment confounders may be unavailable. In this work, 
we propose two strategies for efficiently selecting missing values in pre-treatment confounders to acquire which consist of: i) explicitly characterizing amount of imbalance between treated and control groups and ii) relying on estimation errors of outcome estimates.
We observe benefits of relying on statistical errors versus imbalance in acquisition strategies, and provide insights on their explore-exploit properties and sample choice based on the complexity of outcome response functions. 
Our results highlight the importance of addressing the trade-off between estimation bias and covariate imbalance for this task. Other datasets providing ground truth of the treatment effect could further validate our proposed methods, which we leave for future work.





\acks{Resources used in preparing this research were provided, in part, by the Province of Ontario, the Government of Canada through CIFAR, and companies sponsoring the Vector Institute\footnote{\url{www.vectorinstitute.ai/partners}}. We also thank Bret Nestor and Sindhu Gowda for their suggestions and comments.}

\bibliography{jmlr-sample}

\newpage
\appendix

\newpage
\section{Simulation Algorithms}\label{app:sim_algo}

\begin{algorithm}[htbp]
    \caption{Algorithm to generate missing $A$ values}
    \label{alg:algo_a}
    {\begin{flushleft}
    For each data point $i$
    \begin{itemize}
        \item Sample $u_i$ from $\mathcal{N}((0, 1)$
        \item Compute $p_i = (2 - A_i) \times 0.2 + u_i \times 0.5$
    \end{itemize}
    Rank data point based in descending order of $p_i$ and mask $A$ values for the top $95\%$.
    \end{flushleft}}
\end{algorithm}

\begin{algorithm}[htbp]
\floatconts
    {alg:algo_treatments}
    {\caption{Algorithm to generate the treatments}}
    {\begin{flushleft}
    \textbf{Input}: a subset of features from $X$ ($X_{sub}$), and $\xi$ ($\xi \in \mathbb{R}^n$ where n is the number of features in $X_{}sub$) \\
    \end{flushleft}
    {Compute $p = Clip( X_{sub}\xi^T, 0.005, 0.995)$}
    {Sample $t$ following $Bernoulli(p)$}
    \begin{flushleft}
    \textbf{Output}: $t$. \\
    \end{flushleft}}
\end{algorithm}

\begin{algorithm}[htbp]
    \caption{Algorithm to generate the outcomes}
    \label{alg:algo_outcomes}
    {\begin{flushleft}
    \textbf{Input}: $X$, and  $\beta$, and $W$. ($\beta, W \in \mathbb{R}^n$ where n is the number of covariates.) \\
    \end{flushleft}
    {Compute $\beta$ and $W$.}
    \begin{itemize}[leftmargin=*]
        \item $\beta$
        \begin{itemize}[leftmargin=*]
            \item {Specified values for $b.marr$, $mom.scoll$, $work.dur$, $momwhite$, $cig$, $drugs$ }
            \item {For other continuous covariates: generate a vector of coefficients [0, 0.1, 0.2, 0.3, 0.4], sampled with probabilities [0.5, 0.125, 0.125, 0.125, 0.125].}
            \item {For other binary covariates: generate a vector of coefficients [0, 0.1, 0.2, 0.3, 0.4], sampled with probabilities [0.6, 0.1, 0.1, 0.1, 0.1].}
        \end{itemize}
        \item {$W$} 
        0 for the 6 features with specified $\beta$ values; 0.5 for other covariates 
    \end{itemize}
    {Sample $y_{0}$ following $\mathcal{N}(\exp{((X+W)\beta^T)}, 1)$} \\
    {Sample $y_{1}$ following $\mathcal{N}((X+W)\beta^T, 1)$}
    \begin{flushleft}
    \textbf{Output}: $y_{0}$, $y_{1}$. \\
    \end{flushleft}}
\end{algorithm}

\newpage
\onecolumn
\section{PEHE Results for Section \ref{sec:a_exp}} \label{app:a_exp_pehe}

\begin{table*}[!htb]
\setlength{\tabcolsep}{4pt} 
\small
\floatconts
 {tab:a_exp_pehe}
 {\caption{Speed of reaching optimal PEHE across acquisition strategies while relaxing the independence assumption. 
 Values are the average number of samples needed in training to reach within 1\% of the optimal performance level, with $\pm$ showing the width of 95\% confidence interval, over 100 realizations of the simulated data. Lower is better. Statistically significant better performing strategies for each experiment type are bolded.}}
 {%
   \subtable[Experiments involving original $A$ values. Columns represent fraction of original $A$ values remaining.]{%
    \label{tab:a_exp_original_a_pehe}%
    \begin{tabular}{|l|r|r|r|r|r|r|}
    \toprule
                         & 0              & 0.2            & 0.4                           & 0.6            & 0.8             & 1              \\
    \midrule
    Random               & 436 $\pm$ 28         & 439 $\pm$ 25         & 439 $\pm$ 26                        & 425 $\pm$ 27         & 413 $\pm$ 28          & 392 $\pm$ 34        \\
    Uncertainty          & 420 $\pm$ 31         & 413 $\pm$ 27         & 389 $\pm$ 27                        & 389 $\pm$ 31        & 342 $\pm$ 30          & 341 $\pm$ 32         \\
    CB                   & 396 $\pm$ 28         & 357 $\pm$ 28        & 410 $\pm$ 28                      & 379 $\pm$ 28         & 395 $\pm$ 29          & 406 $\pm$ 30         \\
    OE                   & \textbf{160 $\pm$ 26} & \textbf{159 $\pm$ 24} & \textbf{151 $\pm$ 23}                & \textbf{163 $\pm$ 27} & \textbf{161 $\pm$ 29} & \textbf{152 $\pm$ 25} \\
    \bottomrule
    \end{tabular}
   }\\
   \subtable[Experiments involving bivariate Gaussian simulation. Columns represent correlation coefficient between simulated $A$ and $birthweight$.]{%
    \label{tab:a_exp_birthweight_a_pehe}%
    \begin{tabular}{|l|r|r|r|r|r|r|}
    \toprule
                         & 0              & 0.2            & 0.4                           & 0.6            & 0.8             & 1              \\
    \midrule
    Random               & 453 $\pm$ 26         & 457 $\pm$ 22         & 444 $\pm$ 28                        & 422 $\pm$ 27         & 439 $\pm$ 27          & 431 $\pm$ 26         \\
    Uncertainty          & 421 $\pm$ 32         & 432 $\pm$ 30         & 434 $\pm$ 29	                       & 434 $\pm$ 32         & 456 $\pm$ 24          & 460 $\pm$ 23        \\
    CB                   & 372 $\pm$ 30         & 424 $\pm$ 27         & 440 $\pm$ 25                       & 421 $\pm$ 27         & 460 $\pm$ 23          & 459 $\pm$ 23         \\
    OE                   & \textbf{179 $\pm$ 27} & \textbf{168 $\pm$ 29} & \textbf{164 $\pm$ 25}                & \textbf{181 $\pm$ 30} & \textbf{178 $\pm$ 28} & \textbf{158 $\pm$ 25} \\
    \bottomrule
    \end{tabular}
   }
 }
\end{table*}


\end{document}